# The Future of Artificial Intelligence (AI) and Machine Learning (ML) in Landscape Design: A Case Study in Coastal Virginia, USA

Zihao Zhang[1], Ben Bowes[2]

[1]University of Virginia, Virginia/USA · zz3ub@virginia.edu
[2]University of Virginia, Virginia/USA · bdb3m@virginia.edu

**Abstract:** There have been theory-based endeavours that directly engage with AI and ML in the landscape discipline. By presenting a case that uses machine learning techniques to predict variables in a coastal environment, this paper provides empirical evidence of the forthcoming cybernetic environment, in which designers are conceptualized not as authors but as choreographers, catalyst agents, and conductors among many other intelligent agents. Drawing ideas from posthumanism, this paper argues that, to truly understand the cybernetic environment, we have to take on posthumanist ethics and overcome human exceptionalism.

## 1 Introduction

In 2000 ecologist Robert Cook raised a critical question to the profession of landscape architecture by asking "do landscapes learn?" His concern was the development and adaptation of a landscape design after its construction (COOK 2000). In a way, the landscape theory and practice of the past two decades has been attempting to respond to this question. Based on the new paradigm of ecological research that has moved away from the deterministic and reductionist understanding of nature as a static entity and embraced complexity, dynamism, and probability (REED & LISTER 2014a), the landscape discipline has applied ecology as a powerful tool to design for indeterminacy and emergence (WALDHEIM 2006). For example, in the entry for the Downsview Park competition in late 1990s, James Corner, Stan Allen and Nina-Marie Lister proposed a framework of processes and strategies that could allow for emergent ecologies so that the natural systems can evolve over time and become more complex (CZERNIAK 2001, REED & LISTER 2014b). In other words, to learn means to develop new relationships between different things. The landscape discipline has developed a kind of ecological design paradigm with the notion that nature is not a static entity, but a set of relationships that are constantly changing. The designers' role is to cultivate new relationships in the natural world through landscape design. Projects like Downsview Park serve as exemplars for this new design paradigm that relies on process-based strategies to organize the patterns of material, energy, and information in the environment.

Deeply rooted in this ecological design paradigm, advances in cybernetic technologies such as sensing networks, robotics, and artificial intelligence provide new possibilities to design "cyborg ecologies". In a way, the old notion of "learning" takes on a new meaning. Landscapes cannot only "learn" through ecosystems, but also through coupled technical-natural systems. In this line of research, CANTRELL & HOLTZMAN (2016) have explored how responsive technologies can be used in landscape design so that the coupled system could better respond to various inputs. ZHANG (2017) introduced human factors and the role of socio-cultural systems in the responsive landscape design framework. ESTRADA (2018) imagined sensing networks, data processing and actuating systems in a fluvial environment, and through rigorous experiments, the coupled technical-natural system shows a level of real-time response that is beyond human capability. A responsive landscape does not directly engage ML and AI, but the sensing-processing-actuating feedback loop can serve as a robust framework for the application of AI and ML in landscape design.

There have been theory-based endeavours that directly engage with AI and ML in the landscape discipline. In a thought experiment, CANTRELL et al. (2017) have imagined a DRL (deep reinforcement learning) based AI called "wildness creator" that can devise strategies beyond human comprehension, in order to challenge environmental designers, conservationists, and environmental engineers to reflect on what it means to construct wild places. CANTRELL & ZHANG (2018) have formed machine intelligence as "a third intelligence" in landscape media that can interact with biological intelligence and material intelligence and co-evolve with other landscape agents to co-produce landscapes beyond designers' intention.

The discussion of AI and ML in the landscape architecture discipline remains largely theoretical and speculative. There is a body of practice-based research outside the discipline. For example, Topos, an urbanism-oriented AI start-up has developed ML applications that analyse urban patterns.[1] However, the way how AI and ML is approached is not compatible with the landscape's ecological paradigm. In other words, the complexity of the natural world and ecosystem dynamic are not within their major concern.

To mitigate the gap between theory and practice, this paper presents a case study that directly engages ML and AI in the landscape context in order to raise new challenges for theory building to incorporate ML and AI in the ecological design framework. This paper is organized into three sections. In section one we introduce the key terms such as ML, AI, deep learning, as well as their differences, given the assumption that this field of research is still largely outside the profession of landscape architecture. In section two, we present an interdisciplinary research project that looks into possibilities for the application of machine learning techniques in managing flooding in the coastal city of Norfolk, Virginia, USA. By focusing on the "nuts and bolts" in ML, we present one specific case where ML is used to predict the groundwater table level during storm events – a type of hydrological pattern that is intrinsically related to flooding issues in coastal cities. In section three, we offer a critical analysis of the challenges posed by the case study. Designer's agency, creativity, and intentionality are at stake in an environment that is laden with nonhuman intelligence including AI and other types of intelligence. In this paper, we argue that to truly understand the cybernetic environment that we are moving into, we have to take on a posthumanist ethic and overcome human exceptionalism.

## 2    Machine Learning (ML) and Artificial Intelligence (AI)

The past few years have seen a major development in the research of artificial intelligence (AI), especially in the area of machine learning (ML) and deep learning, which largely revolves around artificial neural networks (ANN). Neural networks are often used to approximate nonlinear functions, advances in computing power have allowed neural networks to become "deep", by increasing the number of neural network layers in a model, which allows them to learn and more accurately represent complex functions. In this context, machines have demonstrated promising results and sometimes out-performed humans in areas such as image recognition (RUSSAKOVSKY et al. 2015), natural language processing and translation (YOUNG et al. 2017), video games (MNIH et al. 2013, 'OpenAI Five' n.d.), and the game of GO (SILVER et al. 2017).

The goal of this section is to familiarize the readers with the capacities as well as limitations of current ML techniques using simple language, given our consideration that the audiences of this paper are largely outside the AI research community. We also want to point to some caveats and clarify the differences between AI and ML, which are almost always used interchangeably by non-experts.

The definition of "universal intelligence" proposed by computer scientists Shane Legg and Marcus Hutter is conceptually stimulating to clarify the difference between AI and ML. Generally speaking, "[i]ntelligence measures an agent's ability to achieve goals in a wide range of environments" (LEGG & HUTTER 2007). This definition relies on a cybernetic model by imagining an intelligent system or agent that interacts with its environment and achieves goals through feedback loops. In order to better achieve goals, this system has to have a representation or model of the environment. ML is one way to help the computer to build and improve this internal representation of the environment by iterating through large amounts of data. AI measures the quality of this representation in terms of helping the agent to achieve goals in the environment.

Partly because of science fiction and pop culture in recent decades, people now tend to use AI to signify another term – artificial general intelligence. Using our definition of universal intelligence, AGI refers to the ability an intelligent agent has compared to a human level of intelligence, like a robot in a sci-fi movie. However, current research in AI is far from building an intelligent machine that can outperform humans in all aspects. Even the AlphaGo, which beat the best human GO player in 2016, is only able to perform well in a very specific area of human culture. Another caveat we want to point out is that, from the definition of intelligence, AI only refers to the ability that a system possesses rather than directly to the system or the agent itself. In other words, AI is a property of a system rather than the system itself. However, in everyday use, people tend to refer to something as an AI system rather than the property of something ("Siri is Apple's AI"). But we do acknowledge the change in meanings of terms, just like the fact that "computer" used to be a job title rather than a tool we use every day.

## 3　Case Study: Machine Learning and Coastal Environmental Management

Along with advances in AI research is the increased accessibility to tools and technologies that allow non-experts such as designers and environmental engineers to practice ML and develop AI systems. Designers and non-experts could use Python-based ML libraries such as PyTorch, Tensorflow, Caffe, or Scikit-learn[2] to train models to perform a specific task. Designers can also use visual programming languages (VPL) such as Grasshopper, a plugin for the 3D modelling software Rhinoceros, to do ML using plugins such as LunchBoxML[3].

At the University of Virginia, the School of Engineering has established a research incubator called the LinkLab that brings together experts and scholars in systems and environmental engineering, computer sciences, social sciences, and landscape architecture to work together and to explore possibilities to apply sensing networks, ML, and cyber-physical systems in managing various environments and landscapes. Among the projects, dMIST (The Data-driven Management for Interdependent Stormwater and Transportation Systems) is an NSF (National Science Foundation) funded research project that looks into coastal water and transportation systems in the City of Norfolk and the greater Hampton Roads metropolitan area on Virginia's east coast. Researchers use data-driven and ML techniques to develop models that build correlations between stormwater and transportation systems so that the coupled system can be more resilient and adaptive to environmental disturbances such as sea level rise, flooding, and climate change. For example, some researchers have developed algorithms to evaluate the trustworthiness of the personal weather stations in Norfolk based on crowdsourced weather data collected by hobbyists and concerned citizens (CHEN et al. 2018). Others use flood event data to train models to predict traffic patterns in the road networks (FARIA et al. 2018, SHRADDHA et al. 2019). One researcher has studied a model predictive control (MPC) strategy in a simulated stormwater management system, the trained model is able to decide when and to what level to open and close floodgates based on forecasts of future precipitation events (SADLER et al. 2019). Many of the models were built with Tensorflow and Keras[4], and are run on the Rivanna High Performance Computer (HPC)[5] at the University of Virginia using a Graphical Processing Unit (GPU).

To illustrate how ML is used in this research, we present a brief description of a study that explores the use of two kinds of neural networks, Recurrent (RNN) and Long Short-term Memory (LSTM), to train models to predict the groundwater table response to storm events (BOWES et al. 2019). In this case, the RNN models were "shallow" whereas the LSTM models, by the definition of their structure, are considered to be deep. The models were built for each of seven groundwater level monitoring wells in Norfolk with hourly data of tide, rainfall, and groundwater table from 2010-2018. Model training in this case refers to the process of adjusting weights in the network to minimize the error between its predictions and the actual observed groundwater table level. After training, the models can predict the groundwater level using only forecast tide and rainfall data. In order to test the model performance, the dataset is always divided into training data that is used to adjust the neural networks, and testing data, which is used to evaluate the model's performance. The models are blind to the testing data at all times.

Two different datasets were used to evaluate the effect of data preprocessing on model performance, the full dataset consisted of the continuous time series (both storm events and no storm event), while the second dataset consisted of only storm events.

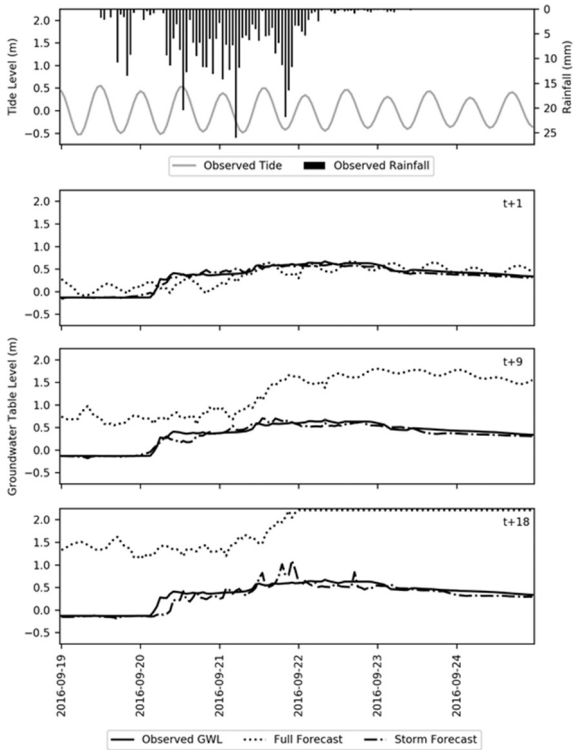

**Fig. 1:**
Comparison of groundwater table forecasts at one well in Norfolk, Virginia from LSTM models trained with the full and storm training sets

Figure 1 shows the testing result of the LSTM models. "Storm Forecast" and "Full Forecast" are two predictions made by models trained with storm event data and full data respectfully, and they are compared to the "Observed GWL", which is the real groundwater level data collected in one of the monitoring wells in Norfolk. This study found that training models with storm event data resulted in better model performance in a simulated real-time forecasting scenario, confirming that a model trained under a certain phenomenon will be more sensitive to this phenomenon.

These models can use forecast rainfall and tide data to create precise estimates of the groundwater table response to storms for every monitoring well in Norfolk, human experts in the system may have a mental model of the storm event-groundwater level relationship, but would not be able to quantify or easily share it with non-experts. Additionally, the precise forecasts of the groundwater table from the ML models can be interpolated into city-wide maps or used to augment flood models. The groundwater table is rarely monitored in coastal cities and decision makers generally have no knowledge of its role in flooding from storm events. Additionally, coastal groundwater table levels will increase with sea level rise and have a growing impact on flood severity. Therefore, understanding the groundwater table patterns and being able to make real-time predictions based on forecast storm conditions of rainfall and tide will help increase the accuracy of real-time flood prediction. Not only will this help governments and the public to better prepare for potential storm events but it will also allow ecologists, environmental designers, and landscape architects to make inferences on the ecological patterns in a coastal environment and make informed design decisions.

# 4     Conclusion: Towards a Cybernetic Environment

The case we have discussed is just one of many emerging efforts in adopting machine learning techniques in managing the environment. Many experiments show that machines can sometimes find patterns in data that humans overlook, and in some cases, they can devise strategies that outperform human strategies. Machines and machine intelligence will be important players in constructing environments in a foreseeable future, and it will be harder and harder to tell when a decision is made by a machine or a human and whether a landscape is created by machine intelligence or human intelligence. For example, if we imagine the City of Norfolk eventually adopted many different models developed by the dMIST group as decision support systems, then human managers' experience of the environmental systems would be highly mediated by these models. Sometimes, processes might be fully automated by intelligent systems, i. e. an MPC system opening and closing the floodgates based on weather forecasts. Moreover, with more and more sensing networks and information technologies embedded in the environment, there will be increased connectivity between different systems, some are technical and some are biological. The new complexity is characterized by numerous numbers of feedback loops between all kinds of intelligent agents including machines, animals, plants, and humans. We argue that "cybernetic environment" can best describe this new paradigm of environmental design. Norbert Wiener in 1948 defined cybernetics as "the scientific study of control and communication in the animal and the machine" (WIENER 2000). Today, the environment of all humans and species has become the new frontier of cybernetic control.

In the cybernetic environment, designer's intentionality, creativity, and agency are at stake since design decisions will inevitably be modified by different systems along the chain of feedback loops in the cybernetic environment. In other words, with the increasing number of biotic and abiotic intelligent agents in the environment being included in the ecological design framework, the designer transfigured from the source of authorship into one active player among many players that shape the environment together. The old design mentality holding that designers are exceptional agents to transform the environment is no longer valid in the cybernetic environment.

We argue that a posthumanist ethic is needed to overcome human exceptionalism in the cybernetic environment. Posthumanism is a transdisciplinary movement that challenges many outdated humanist values and expands the sphere of moral considerations into the more-than-human realm (HAYLES 1999, WOLFE 2010). Posthumanism is not post-human and will never be anti-human, it is rather post-human*ist*. In other words, it challenges a kind of anthropocentric worldview and human exceptionalism. Posthumanism is sympathetic towards many humanist claims but is trying to reveal the presumptions that undercut humanists' moral imperatives. For example, we want to treat animals better not because they are a diminished version of us, but because they are intrinsically different from us. In a similar manner, there is no need to be afraid of machines and no necessity to measure machine intelligence against human intelligence, because they are intrinsically different types of intelligence. Different humans, machines, animals, and plants relate to their outside environment differently, and they achieve their goals relying on totally different perceptive functions and mental processes. By taking on a posthumanist ethics, we can easily conceptualize designers not as authors but as *choreographers, catalyst agents*, and *conductors* among many other intelligent agents. We hope these terms can help to explore a new design paradigm and merit further discussion.

## Acknowledgements

The dMIST project has been supported as part of a National Science Foundation grant: Award #1735587 (CRISP – Critical, Resilient Infrastructure Systems). The authors thank Michael Gorman, Kristina Fauss and other dMIST group members for their support and feedback.

## Footnote Text

[1] https://www.fastcompany.com/90131540/this-startup-used-ai-to-redefine-n-y-c-s-five-boroughs-for-today

[2] Pytorch: https://pytorch.org/; Tensorflow: https://www.tensorflow.org/; Caffe: http://caffe.berkeleyvision.org/ Scikit-learn: https://scikit-learn.org/stable/

[3] https://provingground.io/2017/08/01/machine-learning-with-lunchboxml/

[4] Keras is a library that makes it is easier to build models using Tensorflow: https://keras.io/

[5] https://arcs.virginia.edu/rivanna